\documentclass[twoside,11pt]{article}

\usepackage{blindtext}
\usepackage{graphicx}
\usepackage{amsmath}
\usepackage{caption}
\usepackage{float}
\usepackage[nottoc]{tocbibind}
\usepackage[square,numbers]{natbib}

\usepackage{dmlr2e}

\usepackage{lastpage}
\def\openreview{\url{https://openreview.net/forum?id=RDpDFzijX4}} 

\firstpageno{1}

\begin{document}

\title{Exploring the Efficacy of Meta-Learning: Unveiling Superior Data Diversity Utilization of MAML Over Pre-training}

\author{\name Kavita Selva \email kselva@cs.stanford.edu \\
       \addr Department of Computer Science\\
       Stanford University\\
       Stanford, CA 94305, USA
       \AND
       \name Satita Vittayaareekul \email satita97@stanford.edu \\
       \addr Department of Computer Science\\
       Stanford University\\
       Stanford, CA 94305, USA
       \AND
       \name Brando Miranda \email brando9@stanford.edu \\
       \addr Department of Computer Science\\
       Stanford University\\
       Stanford, CA 94305, USA}

\editor{My editor}

\maketitle
\begin{abstract}
Currently, data and model size dominate the narrative in the training of super-large, powerful models. 
However, there has been a lack of exploration on the effect of other attributes of the training dataset on model performance. 
We hypothesize that dataset \textit{diversity} can impact the performance of vision models. 
Our study shows positive correlations between test set accuracy and data diversity, providing an argument for furthering the research of dataset attributes beyond size.
We analyzed pre-training and model-agnostic meta-learning methods on twelve popular visual datasets (e.g., Omniglot, CIFAR-FS, Aircraft) and five model configurations, including MAML variants with different numbers of inner gradient steps and supervised learning. We show moderate to strong positive correlations (R-squared: 0.15-0.42) between accuracy and data diversity and weaker but significant correlations (R-squared: ~0.2) between loss and diversity.
These findings support our hypothesis and demonstrate a promising way for a deeper exploration of how formal data diversity influences model performance. 
This initial study highlights the potential of (Task2Vec) data diversity as a valuable measure in the rapidly evolving field of large-scale learning and emphasizes that understanding the dataset is key to building more powerful and generalizable models.
\end{abstract}


\section{Introduction}
Current trends in building large, robust vision models emphasize scaling model size and complexity (\cite{chowdhery2022palm}, \cite{nostalgebraist2022chinchilla}, \cite{gpt4}, \cite{google2023palm2}), but other characteristics of datasets, like diversity, are vague and overlooked (\cite{david2010impossibility}, \cite{longpre2023pretrainer}). 
Our study departs from the prevailing focus on data size (\cite{hestness2017deep}, \cite{rosenfeld2019constructive}, \cite{henighan2020scaling}, \cite{kaplan2020scaling}, \cite{gordon2021data}, \cite{hernandez2021scaling}, \cite{jones2021scaling}, \cite{zhai2022scaling}, \cite{hoffmann2022training}, \cite{clark2022unified}, \cite{neumann2022scaling}) and explores this gap by proposing a paradigm shift: encouraging investigation of other dataset characteristics, like diversity, as key ingredients for building more generalizable and performant vision models. 

Our \textbf{contributions} are exploring data diversity as a promising factor in enhancing vision model performance and demonstrating that dataset diversity positively correlates with model efficacy across various model configurations. Moreover, our findings suggest that in the context for meta-learning, the ability to rapidly acquire new knowledge is significantly enhanced by higher data diversity.

\section{Methodology}
We measure training data diversity using the Task2Vec metric (Figure 1) introduced by \cite{miranda2022curse}, which quantifies intrinsic dataset variability in a few-shot learning setting. 
If tasks are considered as probability distributions, this metric provides an approximation of the mean distance between these distributions.
 
The Task2Vec diversity coefficient is formally defined as the expected (cosine) distance observed between the Task2Vec embeddings associated with different tasks or data batches (\cite{miranda2023pretraining}):

\begin{equation}
\widehat{div}(D) = E_{B1, B2\sim D}d(\vec{f}{B1}, \vec{f}{B2})
\end{equation}

where D is the natural language dataset from which we sample batches B1, B2, and \(\vec{f}_{Bi}\) is the Task2Vec embedding of a batch Bi using the diagonal of the FIM matrix. 

\section{Results}

\textbf{Experimental Setup:} We analyze pre-training and model-agnostic meta-learning on twelve popular visual datasets (e.g., Omniglot, CIFAR-FS, Aircraft).
Our experimental setup uses the exact same architecture and minimally adapts the code of \cite{tian-rethinking}.
Performance was measured by test accuracy across configurations.

\begin{figure}[H]
    \centering
    \includegraphics[scale=0.65]{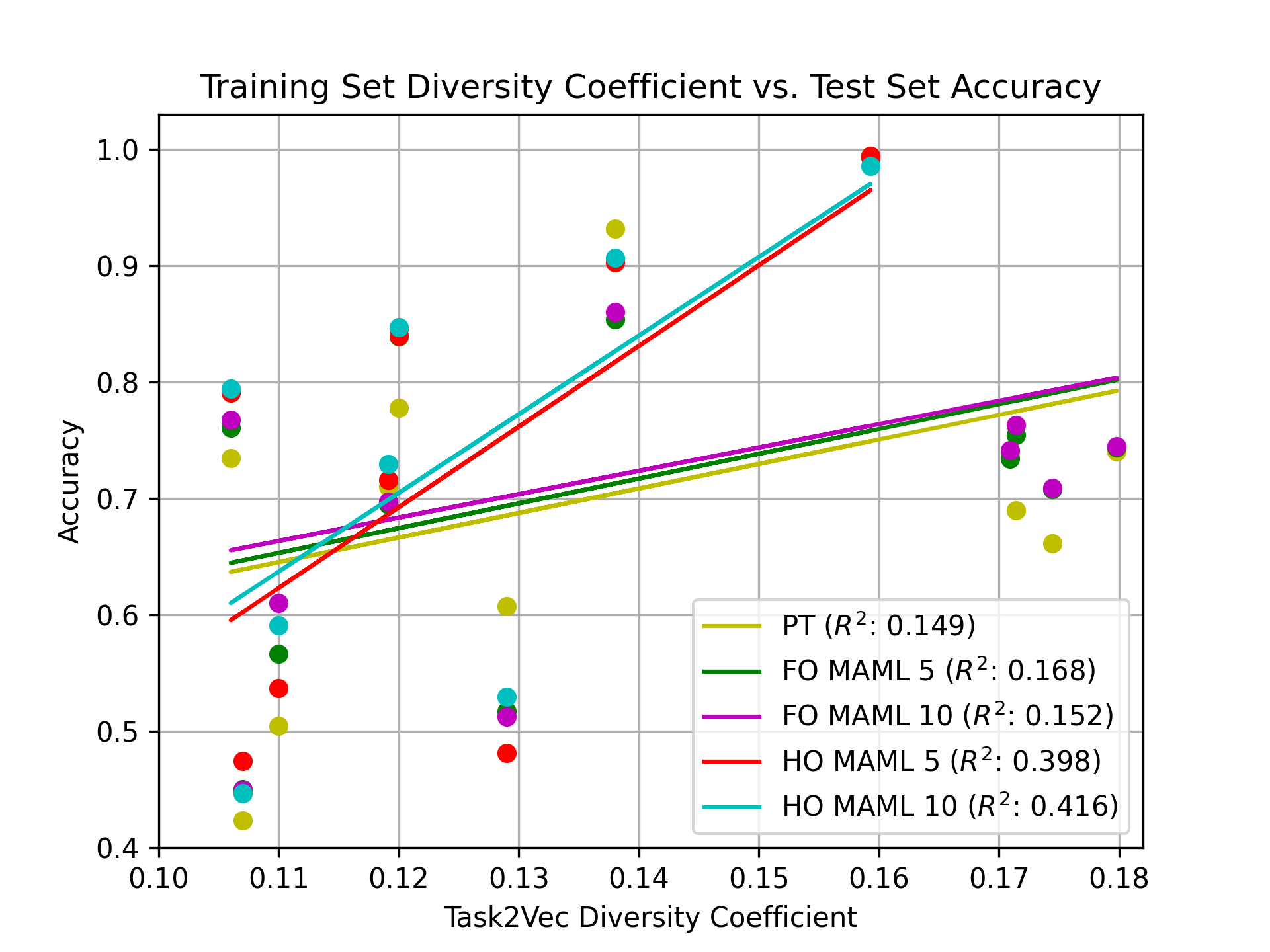}
    \caption{\textbf{We validate that training on more formally diverse data leads to better performance across different settings.}
    We do this by showing moderate to strong positive correlations between accuracy and data diversity in HO MAML models and weaker but significant correlations in FO MAML, PT models. 
    HO MAML stands for Higher-Order Model Agnostic Meta-Learning, FO stands for First-Order Model Agnostic Meta-Learning, and PT for Pre-training.
}
    \label{fig:enter-label}
    \vspace{-15pt}

\end{figure}

Figure \ref{fig:enter-label} depicts the observed relationship for each model. 
Notably, HO (Higher Order) MAML models with 5 and 10 inner gradient steps demonstrate a significantly stronger positive correlation between dataset diversity and relative model performance compared to the PT model and FO (First Order) MAML models. 

This observation is reinforced by the positive R-squared correlation coefficient -- suggesting that data diversity generally improves model performance. 
Significantly higher $R^2$ values were observed for HO MAML 5 and HO MAML 10 models. The HO MAML 10 model exhibits the highest $R^2$ of 0.4 and 0.2, for Accuracy and Cross Entropy loss, respectively (Table \ref{tab:rsquared}).
Similar trends were observed for HO MAML 5. 

\section{Discussion}

We acknowledge a potential confounding factor: an uncontrolled number of data points across training checkpoints. However, the Task2Vec diversity coefficient is designed to capture intrinsic average information invariant to dataset size. This property, coupled with the observed consistent positive correlations between diversity and performance across multiple model configurations (supervised learning, first-order MAML, higher-order MAML), suggests that average diversity or information content is a critical factor influencing model performance. In addition, we observe that when focusing on the overall trend, although the number of data points would provide a more precise analysis, the overall trend of positive correlations between diversity and performance is still informative.
We conjecture that by controlling the number of data points, we'd likely get larger $R^2$ values, but due to expensive training costs, we leave it for future work. 

\section{Conclusion}
Our research unveils the potential of data diversity as a crucial factor in boosting the performance of vision models beyond the current emphasis on scaling model size and complexity. 
Our analysis showed a positive correlation between data diversity and model performance, indicated by a consistent upward trend across all investigated configurations.
Notably, meta-learning models demonstrated a significantly stronger positive correlation with diversity compared to pre-trained approaches. 
Our findings suggest a paradigm shift towards “quality-aware" data selection could address the growing computational costs associated with scaling models and contribute to a more efficient and impactful research trajectory in exploring more powerful AI.

\newpage

\impact{By fostering rigorous data quality analysis, this research empowers both researchers and practitioners to make informed decisions in vision development and from longer-term perspective, LLM development. Societal consequences of this work include reduced costs through the improvement of data selection, which subsequently minimizes required data quantities, lowering development and inference costs and making vision models and more farsighted LLMs more accessible. Another important consequence is the potential environmental gains from reduced energy consumption achieved by more efficient training. Finally, research on data quality through diversity naturally encourages data fairness.
Further exploration is warranted to uncover additional societal and ethical implications.}


\acks{Brando Miranda, acknowledges support by NSF 2046795 and 2205329, the MacArthur Foundation, Stanford HAI, OpenAI and Google Inc.
Brando Miranda acknowledges the Stanford Edge fellowship.
We would also like to acknowledge Xiaolei Shi and Daniel Kolano for their contributions to this work. 
We'd like to thank Sanmi Koyejo for fruitful discussions. \nocite{*}}

\vskip 0.2in
\bibliography{main}

\newpage
\appendix
\section{Related Work}
In the field of investigating the intricate relationship between dataset diversity coefficients and model performance, several notable studies have provided valuable insights. 
\newline
\newline
Is pre-training truly better than meta-learning? \cite{miranda2023pretraining} presents a pre-trained model can outperform standard meta-learning algorithms, MAML, in certain circumstances. They conclude that when the formal diversity of a dataset is low, a pre-trained model beats MAML on average, and when the formal diversity of a dataset is high, MAML beats pre-trained model on average. Both scenarios have a small effect size. The limitations of this paper is that the paper primarily focuses on specific comparisons between pre-training and MAML, the simplest of meta-learning algorithms, which may not encompass the full scope of meta-learning methods. Moreover, the paper provides valuable insights into few-shot learning, it does not explore the models in other fields. In our work, we will explore the computer vision field so that the study would offer a more comprehensive perspective on pre-training and meta-learning.
\newline
\newline
Beyond Scale: the Diversity Coefficient as a Data Quality Metric Demonstrates LLMs are Pre-trained on Formally Diverse Data \cite{lee2023scale} evaluates the Diversity Coefficient can be used as a data quality metric and that the diversity coefficient is dependable for building high-quality and diverse datasets for capable LLMs. The coefficient increases when more datasets are used, the number of latent concepts increases and a richer vocabulary is used. And that the higher the coefficient, the more varied and diverse information LLMs can learn from, and vice versa.
\newline
\newline
The Curse of Low Task Diversity: On the Failure of Transfer Learning to Outperform MAML and Their Empirical Equivalence \cite{miranda2022curse} proposes the novel \textit{diversity coefficient} metric, critical to understanding how to measure the formal diversity of a dataset (or few-shot learning benchmark). This benchmark, based off of the Task2Vec method, quantifies the intrinsic diversity/variability of a dataset/tasks in a few-shot learning benchmark. This work proposes and evaluates the coefficient to define the diversity of MiniImagenet and Cifar-fs, but this does not address our project's objective, which is to understand the relationship between dataset diversity and downstream evaluation. We hope to rigorouPTy extend this work in our project by evaluating the dataset diversity and corresponding performance of various models. 
\newline 
\newline
Task2Vec: Task Embedding for Meta-Learning \cite{achille2019task2vec} provides background on Task2Vec, the embedding method that the formal diversity measure we use in our work is based off of. Task2Vec embeds a task by feeding the data through a pre-trained CNN, and computing the diagonal Fisher Information Matrix. This work tests Task2Vec on a large collection of tasks and models, analyzing the embeddings and concluding that they match human semantic intuitions and taxonomical relations between different tasks. It preceeds the Task2Vec diversity measure proposed by \cite{miranda2022curse} to help us further understand the relationship between the dataset diversity and downstream evaluation performance. 
\newline
\newline
Beyond neural scaling laws: beating power law scaling via data pruning \cite{sorscher2023neural} examines methods for improving the scaling of test error with regards to the size of the input dataset, which have generally been shown to scale as a power law. This scaling requires orders of magnitude more data for small improvements in large models across a variety of disciplines. The work examines methods of dataset pruning to train on examples that will best improve model performance. The results show that results better than power law scaling can be achieved using training data pruning, and presents a novel example pruning metric that performs well on the ImageNet dataset.

\section{Methods}

This study employs a rigorous experimental design to investigate the impact of training set diversity on model performance, while maintaining control over confounding variables. We adopted a uniform network architecture and fixed hyperparameters across all models under examination. In addition, we ensured the optimization techniques and convergence criteria were consistent for each model. These methodological constraints were established to ascertain that any observed differences in performance on the test set were attributable to the variation in the diversity of the training datasets, rather than extraneous factors.
\newline \newline
To assess the influence of dataset diversity on model accuracy, we conducted comparative analyses between two prevalent training methods: Pre-training (PT) and Model-Agnostic Meta-Learning (MAML). The selection of these methods was predicated on their relevance and common application in the domain of machine learning, particularly in tasks requiring the generalization of models to new data.

\section{Dataset and Features}
Our empirical analysis draws on a suite of established visual datasets that are benchmarks in the machine learning community, frequently utilized in the context of few-shot learning and meta-learning paradigms. The datasets include CIFAR-FS, FC100, Mini-ImageNet, Aircraft, Flower, DTD (Describable Textures Dataset), CUBirds (Caltech-UCSD Birds 200), Omniglot, MIO, and a series of datasets designed to embody high degrees of diversity, labeled as hdb7-afto through hdb10-micova. Each dataset is carefully curated to encapsulate a unique spectrum of diversity, defined by the variety and depth of latent concepts as well as the richness of the feature space present within the data. Such diversity is believed to enrich the learning experience of models by exposing them to a vast array of information.
\newline \newline
To quantify the diversity present within these datasets, we employed the Task2Vec Diversity Coefficient, a metric which reflects the inherent variability of a dataset and enables a normalized comparison between different training sets. The computation of the diversity coefficient was facilitated by Resnet18 and Resnet34 models previously trained on the ImageNet database. This model functions as the computational framework for generating the Fisher Information Matrix (FIM), which is crucial for the derivation of Task2Vec task embeddings that are used in quantifying dataset diversity. The results of this quantitative assessment are concisely presented in Table \ref{tab:diversity}, which illustrates the diversity metrics of the datasets under study and forms the empirical foundation for exploring the correlation between dataset diversity and model efficacy in subsequent tasks.
\newline \newline
In the data preprocessing phase, all images were resized to a uniform resolution to ensure consistency. Furthermore, pixel values were normalized to a standard scale to mitigate discrepancies in lighting and contrast. These preprocessing steps are vitally important for reducing potential biases that may arise from variations in image quality and format.
\newline \newline
The Feature Information Matrix was obtained using Resnet18 and Resnet34 models that had undergone pre-training on the ImageNet dataset. This pre-trained model provided the necessary foundation for calculating task embeddings via Task2Vec, which in turn facilitated the computation of the diversity coefficient for each dataset. The diversity coefficient is a pivotal element in our analysis, providing a numerical representation of the variation within our datasets, as summarized in Table \ref{tab:diversity}. The table encapsulates the diversity scores for each dataset based on the task embeddings derived from the pre-trained Resnet18 and Resnet34 models, offering a comprehensive view of the datasets' intrinsic variability for subsequent analytical scrutiny.

\begin{table}[ht]
\centering
\begin{tabular}{c|c|c}
\hline
Dataset & Diversity(Resnet18) & Diversity(Resnet34) \\
\hline
CIFAR-FS & $0.106\pm 0.00166$ & $0.0890 \pm 0.00199$ \\
\hline
FC100 & $0.107\pm0.00149$ & $0.0903 \pm 0.00389$ \\
\hline
Aircraft & $0.110 \pm 0.00127$ & $0.0932 \pm 0.00109$ \\
\hline
Flower & $0.138 \pm 0.00288$ & $0.117 \pm 0.00234$ \\
\hline
DTD & $0.129 \pm 0.00227$ & $0.111 \pm 0.00228$ \\
\hline
CUBirds & $0.120 \pm 0.00161$ & $0.104 \pm 0.00149$ \\
\hline
Omniglot & $0.159\pm 0.00472$ & $0.136 \pm 0.00421$ \\
\hline
MIO & $0.1191 \pm 0.0015$ & $0.1013 \pm 0.00122$ \\
\hline
hdb7-afto & $0.171 \pm 0.01435$ & $0.148 \pm 0.01179$ \\
\hline
hdb8-cado & $0.174 \pm 0.01$ & $0.155 \pm 0.00845$ \\
\hline
hdb9-cavdo & $0.180 \pm 0.00920$ & $0.143 \pm 0.00844$ \\
\hline
hdb10-micova & $0.171 \pm 0.0101$ & $0.144 \pm 0.00766$ \\
\hline
\end{tabular}
\caption{Dataset diversities as measured by the Task2Vec diversity coefficient, utilizing the training split with 95\% confidence intervals. The diversity coefficient is calculated using the Fisher Information Matrix (FIM) from the Resnet18 model and Resnet32 model pre-trained (SL) on ImageNet. These embeddings provide a basis for the Diversity Coefficient, reflecting the intrinsic variability of the datasets.}
\label{tab:diversity}
\end{table}

\section{Experiments, Results and Discussion}
\subsection{Experiments}
We assessed the diversity coefficient of 12 publicly available Large Language Model (LLM) pre-training datasets in this study. The diversity coefficient Our analysis aimed to measure the variety and heterogeneity present in these datasets, giving insights into the wide range of linguistic contexts they contain. The datasets we used to evaluate are CIFAR-FS, FC100, Aircraft, Flower, DTD, CUBirds, Omniglot, MIO, hdb7-afto, hdb8-cado, hdb9-cavdo, hdb10-micova. 
\newline \newline
We followed the method of Task2Vec diversity coefficient introduced by Miranda et al.\cite{miranda2022curse} to calculate the diversity coefficient for each datasets we evaluated. The method serves as a quantitative measure to estimate the effective task diversity within a dataset. If tasks are considered as probability distributions, this metric provides an approximation of the mean distance between these distributions. Diversity Coefficient equation as follows:

\begin{center}
    \(\widehat{div}(D) = E_{B1, B2\sim D}d(\vec{f}_{B1}, \vec{f}_{B2})\)
\end{center}
where D is the natural language dataset from which we
sample batches B1, B2, and \(\vec{f}_{Bi}\) is the Task2Vec embedding of a batch Bi using the diagonal of the FIM matrix \(\widehat{F}_{Bi}\)\cite{miranda2023pretraining}.
\newline \newline
The validation process involves synthetic experiments where the ground truth diversity is predetermined and known. Specifically, the Task2Vec diversity coefficient is formally defined as the expected (cosine) distance observed between the Task2Vec embeddings associated with different tasks or data batches \cite{miranda2023pretraining}. This calculation is performed with respect to a consistent probe network in the context of a few-shot learning benchmark or dataset. In essence, the metric captures the relational dissimilarity between tasks, contributing to a more nuanced understanding of the diverse nature of tasks within the dataset.
\newline \newline
In our experimental setup, we maintained consistency in hyperparameters to isolate the impact of data variations on model performance. Specifically, we used the fixed architecture of resnet12 across all experiments. For pre-trained model, only the last layer is fine tuned. To explore the nuances of meta-learning, Model-Agnostic Meta-Learning (MAML) under different configurations is used. MAML was executed in both 1st and 2nd order to capture the intricacies of different meta-learning methodologies. The inner gradient steps is chosen as 5 and 10 for each order MAML. As for the hyperparameters we maintained, the MAML outer learning rate (lr) was set to 1e-3, and inner learning rate is set to 1e-1.The query set size, which represents the number of samples used for testing the model's generalization on unseen data, was configured to 15. The support set size was set to 5 to specify the number of samples utilized for training and facilitating the model's adaptation to the task at hand. The number of ways, denotes the number of classes present in each meta-learning task, was configured to 5.

\subsection{Results}
In all configurations of supervised learning, first order MAML, and higher order MAML, we saw a positive relationship between the diversity coefficient of the dataset and the downstream model performance, measured by test accuracy. In figure 1, we can see the diversity coefficient versus accuracy for the three model settings we explored in this work. In the first order and higher order MAML models with inner gradient steps 5 and 10, we see a stronger relationship between the diversity of the training dataset and relative model performance. We see a weaker relationship with regard to the supervised learning model, and both first order MAML models with inner gradient steps of 5 and 10.
\newline \newline
We can also more rigorously quantitatively analyze our results using the $R^2$ correlation coefficient, which explains the proportion of variance in the response variable (accuracy) that can be explained by the predictor variable (dataset diversity). The $R^2$ results are as follows:

\begin{table}[H]
\centering
\begin{tabular}{c|c|c}
\hline
Model & $R^2$ (Acc) & $R^2$ (CELoss) \\
\hline
PT & $0.149$ & $0.137$ \\
\hline
FO MAML 5 & $0.168$ & $0.184$ \\
\hline
FO MAML 10 & $0.152$ & $0.174$ \\
\hline
HO MAML 5 & $0.398$ & $0.074$ \\
\hline
HO MAML 10 & $0.416$ & $0.203$ \\
\end{tabular}

\caption{$R^2$ correlation coefficients for each model configuration, where the response variable is accuracy or cross entropy loss, and the predictor variable is the dataset diversity.}

\label{tab:rsquared}
\end{table}
\noindent
Interestingly, we can confirm the qualitative results we noticed from figure 1. The $R^2$ coefficient for accuracy and data diversity is higher in the HO MAML 5 and HO MAML 10 models. In particular, in the HO MAML 10 model, we can see that 40 percent of the variance in the accuracy can be explained by the diversity of the dataset, and 20 percent of the variance in the cross entropy loss can be explained by the diversity of the dataset. We also see a similar strength for how well the regression model fits in the HO MAML 5 model for the accuracy. We can make similar but weaker claims for the other models that were investigated in this study. We can easily see a more qualitative understanding of this claim demonstrated in figure 2, which plots each model's trendline with regard to diversity coefficient versus accuracy.

\subsection{Discussion}
From these results, we can make some positive claims (strength depending on configuration) that the diversity of the training dataset has an effect on the downstream performance of the model, bolstered by the $R^2$ analysis. This is an interesting finding as it suggests that understanding and improving the quality of pre-training data can serve to be a useful metric when attempting to improve the performance of vision models, as opposed to simply the more well-known and more expensive method of improving performance (model size and scaling). We hope to continue this discussion by in the future understanding the relationship between the cross entropy loss and the diversity of the training dataset. 

\section{Future Work}
We hope that this research jumpstarts a wider interest in exploring more nuanced methods of improving model performance, especially as current trends of focusing on size and scaling of models become more and more computationally expensive. With results that highlight the importance of dataset quality in terms of making improvements to downstream model performance, we aim to contribute to the conversation of intentionality in improving model performance with not just model configurations, sizes, and dataset scale, but also giving attention to these alternative methods of approaching artificial intelligence questions in general.
\newline \newline 
In the future, we would like to extend our work in multiple ways. First, we would like to consider additional datasets with more variance in their diversity coefficient to strengthen our understanding of its relationship with downstream performance. We would also like to create more plots to analyze the data, specifically diversity versus cross entropy loss, as well as add additional statistical information like confidence intervals.
\newline \newline 
We also wanted to note that there is a potential confounding factor in this work in that the number of data points for each checkpoint was not controlled. In the future, we would like to explore an idea to deal with the different amount of data points by choosing checkpoints with similar average Task2Vec complexity. In other words, this would involve computing the average Task2Vec complexity of each dataset models were trained on, and seeing if they have similar complexities. We would then choose datasets with similar complexity, defining similar potentially as a small effect size difference. 
\newline \newline 
We also recognize that this work explores only performance on vision models. It would be an interesting extension of this work to evaluate how the diversity of the training set affects downstream performance in other applications currently affected by huge scaling of datasets such as language related tasks. Overall, we hope that with our work we have contributed to the grounding of the understanding of methods to getting a performant model through explorations of dataset quality.
\newline \newline 
Moving forward, we aim to conduct more comprehensive experiments across diverse model architectures, methodologies, and datasets to refine our findings and aims to address the gap in exploring the impact of dataset quality on model efficacy, building upon our demonstrated positive correlations between accuracy and data diversity. 
We also propose to select checkpoints with similar average Task2Vec complexity to control for dataset size in future experiments, to isolate the effect of diversity more precisely.

\end{document}